\documentclass{article}
\usepackage{ijcai19}

\usepackage{amssymb} 
\usepackage{tabularx}
\usepackage{balance}
\usepackage{natbib}

\usepackage{times}
\usepackage{soul}
\usepackage{url}
\usepackage[hidelinks]{hyperref}
\usepackage[utf8]{inputenc}
\usepackage[small]{caption}
\usepackage{graphicx}
\usepackage{amsmath}
\usepackage{xspace}
\usepackage{xcolor} 

\usepackage{booktabs}

\usepackage[table]{xcolor}  

\usepackage{makecell}

\definecolor{lightred}{RGB}{255,230,230}

\title{EEG-Titans: Long-Horizon Seizure Forecasting via Dual-Branch Attention and Neural Memory}

\author{
Tien-Dat Pham$^1$\and
Xuan-The Tran$^2$\footnote{\textit{Corresponding Author}}\\
\affiliations
$^1$HAI-Smartlink Research Lab, Anchi STE Company, Haiphong, Vietnam\\
$^2$School of Mechanical Engineering, Vietnam Maritime University, Haiphong, Vietnam
\footnote{\textit{Preprint notice:} This work has been submitted for possible publication. Copyright may be transferred without notice, after which this version may no longer be accessible.}
}

\begin{document}

\maketitle

\begin{abstract}
Accurate epileptic seizure prediction from electroencephalography (EEG) remains challenging because pre-ictal dynamics may span long time horizons while clinically relevant signatures can be subtle and transient. Many deep learning models face a persistent trade-off between capturing local spatiotemporal patterns and maintaining informative long-range context when operating on ultra-long sequences. We propose EEG-Titans, a dual-branch architecture that incorporates a modern neural memory mechanism for long-context modeling. The model combines sliding-window attention to capture short-term anomalies with a recurrent memory pathway that summarizes slower, progressive trends over time. 
On the CHB-MIT scalp EEG dataset, evaluated under a chronological hold-out protocol, EEG-Titans achieves 99.46\% average segment-level sensitivity across 18 subjects. We further analyze safety-first operating points on artifact-prone recordings and show that a hierarchical context strategy extending the receptive field for high-noise subjects can markedly reduce false alarms (down to 0.00\,FPR/h in an extreme outlier) without sacrificing sensitivity. These results indicate that memory-augmented long-context modeling can provide robust seizure forecasting under clinically constrained evaluation.
\end{abstract}

\section{Introduction}
\label{sec:intro}

Epilepsy is among the most prevalent chronic neurological disorders, affecting approximately 50 million people worldwide~\cite{who2019report,fiest2017prevalence}. The stochastic and unpredictable nature of seizures not only increases the risk of physical injury but also imposes a persistent psychological burden, substantially reducing patients’ quality of life~\cite{mormann2007seizure}. Accordingly, developing an accurate forecasting system that can issue early warnings during the pre-ictal period (typically 5--30 minutes before seizure onset) is widely regarded as a key clinical objective in epilepsy management~\cite{mormann2007seizure}. Such a system could enable patients to seek a safer environment and support timely interventions, including medication administration, when appropriate~\cite{cook2013prediction}.

Despite decades of progress, forecasting seizures from EEG remains exceptionally challenging. In contrast to the pronounced morphological signatures observed during ictal events, pre-ictal biomarkers are often subtle, non-stationary, and may evolve gradually over extended durations~\cite{truong2018generalised}. This characteristic implies that effective models must capture and integrate information across long time horizons, i.e., long-term dependencies, rather than relying solely on local patterns~\cite{hochreiter1997long}. Over the past decade, deep learning has achieved notable advances in seizure prediction; however, widely used architectures still face fundamental limitations in jointly modeling local discriminative features and long-range temporal context under practical computational constraints. Convolutional neural networks (CNNs) excel at extracting local morphology (e.g., spikes and sharp waves) but are inherently constrained in representing long-term temporal structure due to limited receptive fields~\cite{truong2018generalised}. Recurrent architectures such as LSTMs are designed for sequential modeling, yet in practice they can struggle to preserve informative context over very long sequences, which may lead to degraded performance when pre-ictal dynamics span thousands of time steps~\cite{hochreiter1997long}. Transformer-based self-attention alleviates the global-context bottleneck, but its quadratic complexity $\mathcal{O}(N^2)$ becomes prohibitive for long observation windows and resource-constrained settings such as wearable devices~\cite{vaswani2017attention,hussein2022multi}.

Motivated by these challenges and inspired by Titans, a recent advance in long-context modeling via test-time memory mechanisms~\cite{titans_paper}, we propose EEG-Titans, a specialized framework for seizure prediction. The main contributions of this work are summarized as follows:
\begin{itemize}
    \item \textbf{Dual-branch architecture with neural memory:} We propose a dual-stream design in which one branch emphasizes instantaneous anomalies, while the other incorporates a neural memory module to accumulate and update latent brain states over time. This enables modeling long-term pre-ictal evolution with substantially lower computational overhead than applying full self-attention over long sequences.
    
    \item \textbf{Safety-first, causal evaluation protocol:} We adopt a safety-first perspective that prioritizes high sensitivity and avoids retrospective protocols that may weaken temporal causality. Concretely, we replace commonly used leave-one-out cross-validation with a chronological hold-out strategy that preserves temporal order between training and testing data.
    
    \item \textbf{Transparent outlier and failure-mode analysis:} Beyond reporting peak performance, we provide an explicit analysis of challenging cases, including artifact-heavy recordings and pediatric subjects. This clarifies the trade-off between long-context memory capacity and noise sensitivity, and motivates future directions toward personalized forecasting systems.
\end{itemize}

\section{Methodology}
\label{sec:methodology}

\subsection{Dataset and Preprocessing}

\subsubsection{Dataset Description and Subject Selection}

This study uses the CHB-MIT Scalp EEG database, a widely used benchmark for seizure prediction research~\cite{shoeb2009application}. The recordings were collected at Boston Children’s Hospital and are publicly distributed through PhysioNet (Massachusetts Institute of Technology). The dataset contains long-term EEG recordings from 24 pediatric subjects (chb01--chb23; chb21 and chb01 correspond to the same individual recorded approximately 1.5 years apart)~\cite{shoeb2009application}. Subjects (5 males, 17 females; age range 1.5--19 years) were diagnosed with intractable epilepsy and underwent inpatient video-EEG monitoring for pre-surgical assessment. Anti-epileptic drugs were reduced or withdrawn during monitoring to increase the likelihood of capturing seizures for clinical evaluation.

CHB-MIT includes over 980 hours of continuous EEG and 198 seizure events annotated by clinical experts~\cite{shoeb2009application}, providing sufficient scale for training models that require extended temporal context. EEG was recorded at 256~Hz with 16-bit resolution, preserving activity across standard physiological bands. Recordings follow the international 10--20 electrode placement system; however, the number of channels varies across subjects (typically 23--26). To ensure a fixed input dimensionality and consistent spatial alignment, we restrict analysis to 18 bipolar channels that are common across all subjects, following established practice in prior work~\cite{truong2018generalised}. Data are provided as \texttt{.edf} files, typically segmented into approximately one-hour recordings, and are freely available for research use via PhysioNet.

To satisfy the requirements of our sequential simulation strategy and maintain strict input consistency, we applied a screening procedure to exclude subjects that do not support reliable chronological evaluation or standardized channel configurations. Specifically, subjects were excluded if their recordings exhibited substantial montage changes or lacked channels required for the fixed 18-channel input; if they did not provide at least two valid seizure clusters necessary for splitting the first $N\!-\!1$ clusters for training and the final cluster for testing; or if their summary files lacked reliable timeline metadata needed to define pre-ictal windows and safety margins.

Applying these criteria, 6 subjects were removed from the original cohort. CHB12 was excluded due to marked montage heterogeneity relative to the standard 10--20 configuration, which would undermine spatial feature alignment. CHB24 was excluded because the timing information required to establish the pre-ictal window and safety margins was not available. In addition, CHB11, CHB17, CHB19, and CHB23 were excluded because they did not contain a sufficient number of seizure clusters to support the proposed chronological train--test split. The final study cohort therefore comprises \textbf{18 subjects}. The sampling rate for all recordings was 256~Hz, which we retain throughout to ensure spectral consistency.

\begin{table}[htbp]
    \centering
    \caption{Statistics of selected CHB-MIT subjects~\cite{shoeb2009application}. Bold values indicate seizures excluded due to artifacts or insufficient pre-ictal margins.}
    \label{tab:chbmit_stats}
    \setlength{\tabcolsep}{6pt}
    \renewcommand{\arraystretch}{1.1}
    \begin{tabular}{lccc}
        \toprule
        \textbf{Subject ID} & \makecell{\textbf{Original}\\\textbf{Seizures}} &
        \makecell{\textbf{Selected}\\\textbf{Seizures}} &
        \makecell{\textbf{Duration}\\\textbf{(Hours)}} \\
        \midrule
        CHB01 & 7  & 7            & 40.55 \\
        CHB02 & 3  & 3            & 35.30 \\
        CHB03 & 7  & 7            & 38.00 \\
        CHB04 & 4  & 4            & 155.90 \\
        CHB05 & 5  & 5            & 39.00 \\
        CHB06 & 10 & 10           & 66.70 \\
        CHB07 & 3  & 3            & 68.10 \\
        CHB08 & 5  & 5            & 20.00 \\
        CHB09 & 4  & 4            & 67.80 \\
        CHB10 & 7  & 7            & 50.00 \\
        CHB13 & 12 & \textbf{11}  & 33.00 \\
        CHB14 & 8  & 8            & 26.00 \\
        CHB15 & 20 & \textbf{17}  & 40.00 \\
        CHB16 & 10 & \textbf{9}   & 19.00 \\
        CHB18 & 6  & 6            & 36.00 \\
        CHB20 & 8  & 8            & 29.00 \\
        CHB21 & 4  & 4            & 33.00 \\
        CHB22 & 3  & 3            & 31.00 \\
        \bottomrule
    \end{tabular}
\end{table}

\subsubsection{Seizure Clustering and Labeling Strategy}
\label{sec:labeling_strategy}

\begin{figure*}[h!]
    \centering
    \includegraphics[width=0.85\linewidth]{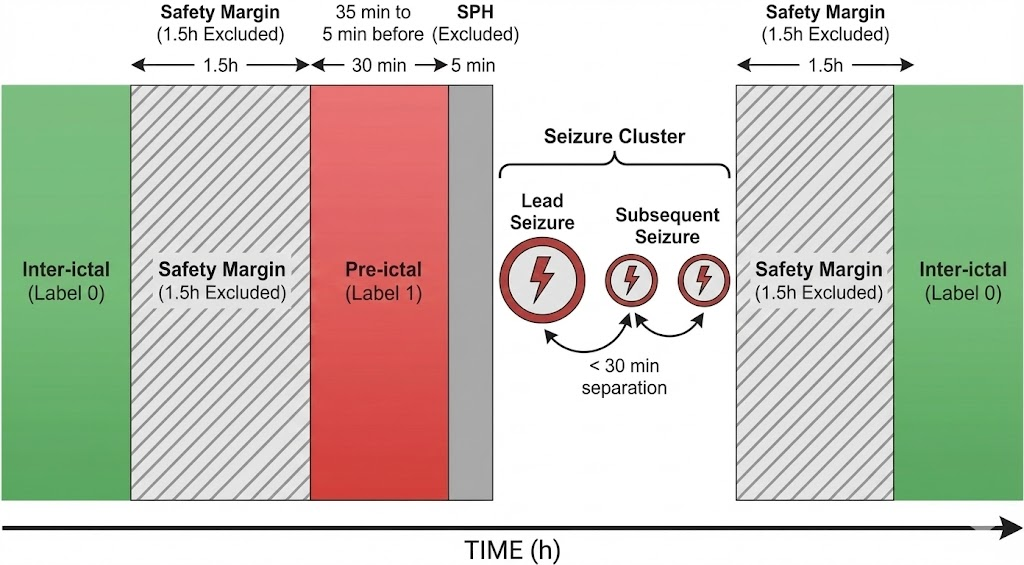}
    \caption{\textbf{Schematic of the seizure prediction protocol.} The pre-ictal interval (Label~1) is defined as a 30-minute window prior to seizure onset and is separated from onset by a 5-minute Seizure Prediction Horizon (SPH). A 1.5-hour safety margin is removed between inter-ictal (Label~0) and seizure-related periods to reduce label leakage and contamination.}
\label{fig:labeling_strategy}
\end{figure*}

As illustrated in Fig.~\ref{fig:labeling_strategy}, seizures in long-term monitoring often occur in temporal clusters. Forecasting seizures that occur shortly after a preceding event offers limited practical value and can confound labeling due to overlapping peri-ictal effects. Following the seizure clustering strategy in prior work~\cite{truong2018generalised}, we group seizures separated by less than 30 minutes into a single cluster, and define the onset of the lead seizure as the reference time for label construction.

Specifically, the pre-ictal phase (Label~1) is defined as the interval from 35 to 5 minutes prior to the onset of the lead seizure (Fig.~\ref{fig:labeling_strategy}). The final 5 minutes before onset correspond to the Seizure Prediction Horizon (SPH)~\cite{mormann2007seizure} and are excluded from training to emulate a clinically realistic buffer time for warning and response. The inter-ictal phase (Label~0) represents baseline EEG far from seizure activity. To reduce contamination by post-ictal effects and potential early physiological drifts near seizure onset, we enforce a safety margin of 1.5 hours around each seizure cluster~\cite{bandarabadi2015epileptic}. Specifically, all EEG within 1.5 hours before and after a cluster is discarded, yielding a conservative inter-ictal set intended to minimize label leakage.

\subsubsection{Signal Processing and Data Augmentation}
\label{sec:signal_processing}

Raw EEG is notch-filtered at 60~Hz to attenuate power-line interference, and then band-pass filtered from 0.5 to 100~Hz to retain physiologically relevant activity. Continuous recordings are segmented into 5-second windows (1280 samples). Given the pronounced class imbalance in seizure forecasting, we adopt a differential windowing strategy inspired by prior work~\cite{tsiouris2018long}. Inter-ictal segments are extracted using non-overlapping windows to limit redundancy and computational overhead. For pre-ictal segments, we apply a targeted overlap strategy during training, using 50\% overlap between adjacent windows to increase the number of positive samples and improve exposure to rare pre-ictal patterns. In validation and testing, we use strictly non-overlapping windows for both classes to ensure a conservative and realistic evaluation and to prevent artificially inflated performance due to near-duplicate samples.

\subsection{EEG-Titans Architecture}
\label{sec:architecture}
\begin{figure*}[htbp] 
    \centering
    \includegraphics[width=0.85\linewidth]{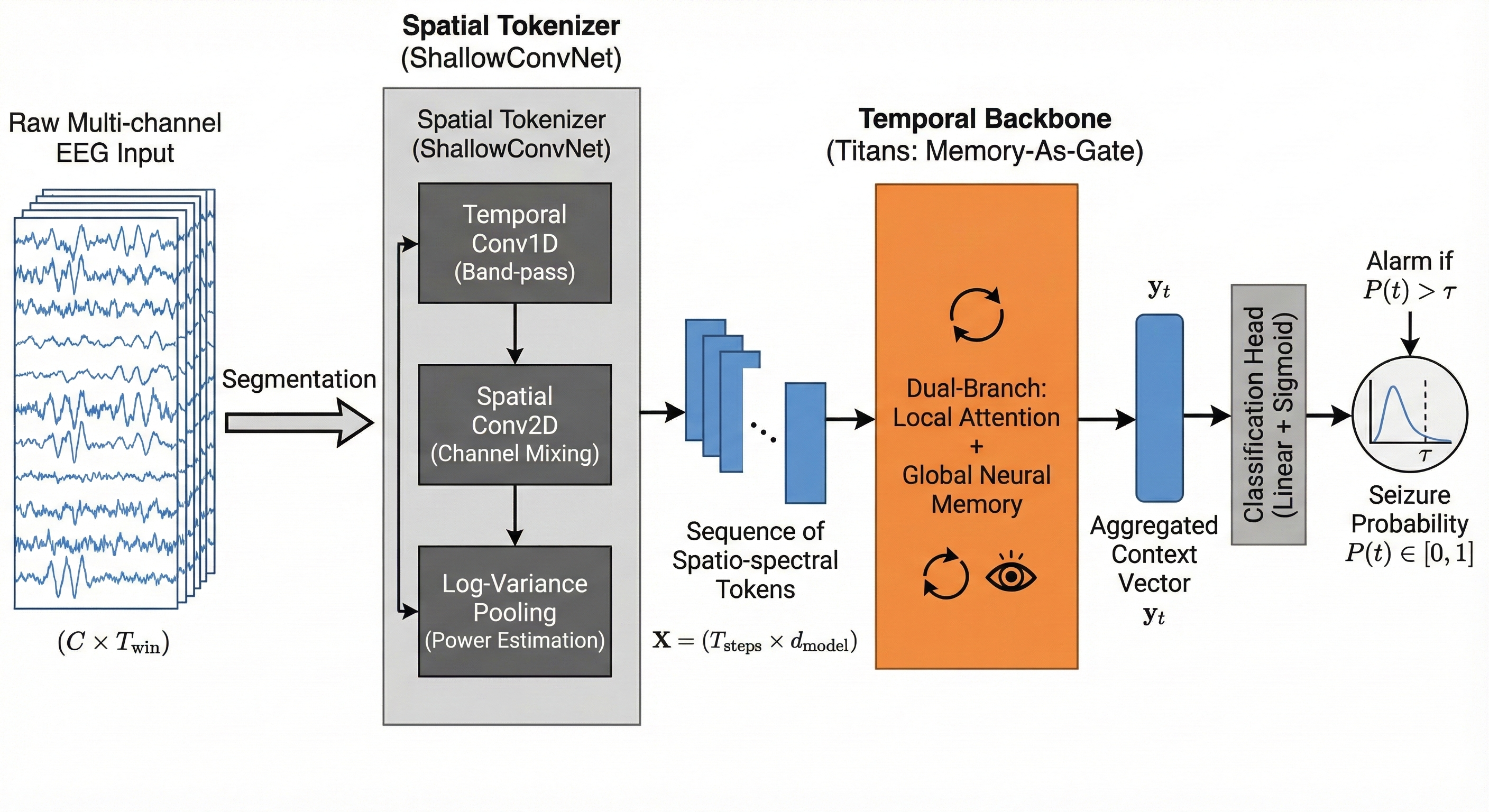}
    \caption{\textbf{Overall architecture of EEG-Titans.} Raw multi-channel EEG is processed by (1) a \textbf{Spatial Tokenizer} based on ShallowConvNet to extract spectro-spatial features and produce a token sequence; (2) a \textbf{Temporal Backbone} based on Titans (Memory-as-a-Gate) to model long-term dependencies via complementary local (attention) and global (memory) pathways; and (3) a classification head that outputs the seizure probability $P(t)$.}
    \label{fig:overall_architecture}
\end{figure*}

This section describes \textbf{EEG-Titans}, a deep learning framework designed for epileptic seizure prediction from multi-channel scalp EEG. As illustrated in Fig.~\ref{fig:overall_architecture}, the model consists of two main stages: a spatial feature extraction and tokenization module, followed by a long-context temporal backbone with neural memory, and a final classification head that outputs a seizure probability.

EEG-Titans operates as a sequential pipeline that maps raw EEG segments to seizure probabilities:
\begin{itemize}
    \item \textbf{Input:} Multi-channel EEG segments are partitioned into fixed-length sub-windows to form the input stream.
    \item \textbf{Spatial Tokenizer (ShallowConvNet):} A task-specific feature extractor that learns spatial and spectral filters and encodes each sub-window into a sequence of feature tokens~\cite{schirrmeister2017deep}.
    \item \textbf{Temporal Backbone (Titans):} A neural-memory-based sequence model that integrates short-range and long-range temporal context over the token sequence~\cite{titans_paper}.
    \item \textbf{Output:} A classification layer produces the predicted seizure probability for each time step (or window) in the sequence.
\end{itemize}

\subsubsection{Spatial Tokenizer - ShallowConvNet}

We adopt ShallowConvNet as the spatial tokenizer because EEG signals are characterized by oscillatory activity distributed across a structured sensor topology~\cite{schirrmeister2017deep}. In contrast to deep CNN backbones originally optimized for static images, ShallowConvNet is explicitly designed to capture spectral--spatial patterns. Conceptually, this architecture can be viewed as a trainable approximation of classical pipelines such as Filter Bank Common Spatial Patterns (FBCSP), enabling end-to-end learning of spatial projections and band-specific activity without manual feature engineering.

\textbf{Technical details:} Let an input EEG segment be $X \in \mathbb{R}^{C \times T}$, where $C$ is the number of channels and $T$ is the number of time samples. The tokenizer is implemented as a sequence of transformations that parallels common EEG processing steps:

\begin{itemize}
    \item \textbf{Temporal convolution:} 
    A temporal convolution extracts frequency-related patterns using a kernel of size $(1, K_t)$ with $K_t=40$. With sampling rate $f_s=256$~Hz, this corresponds to a temporal window of approximately 156~ms. This layer can be interpreted as a learnable temporal filtering stage:
    \begin{equation}
        Z_{\mathrm{temp}} = X \ast W_{\mathrm{temp}} + b_{\mathrm{temp}},
    \end{equation}
    where $W_{\mathrm{temp}}$ denotes the temporal filter weights. The relatively large kernel supports capturing low-frequency rhythms that are often relevant to pre-ictal dynamics~\cite{mormann2007seizure}.

    \item \textbf{Spatial convolution:} 
    The temporal features are then projected across channels using a spatial convolution with kernel size $(C,1)$, producing a set of learned spatial mixtures (virtual channels):
    \begin{equation}
        Z_{\mathrm{spat}}^{(j)} = \sum_{i=1}^{C} Z_{\mathrm{temp}}^{(i)} \cdot W_{\mathrm{spat}}^{(i,j)}.
    \end{equation}
    This operation is analogous to learning spatial projection matrices as in CSP-based approaches, emphasizing informative channel combinations while suppressing irrelevant activity~\cite{schirrmeister2017deep}.

    \item \textbf{Log-variance activation (square and log-pooling):}
    Unlike conventional CNNs that use ReLU, ShallowConvNet employs a nonlinearity designed to approximate signal power. The sequence includes squaring, average pooling over a window of length $W$, and logarithmic compression:
    \begin{equation}
        f(x) = \log\left(\frac{1}{W}\sum_{k=1}^{W} x_k^2\right).
    \end{equation}
    This yields a log-power representation, which is closely related to spectral power features commonly used in EEG analysis and seizure prediction~\cite{bandarabadi2015epileptic}. The logarithm also stabilizes variance and mitigates dynamic-range effects.

    \item \textbf{Tokenization:}
    The resulting features are projected to the model dimension $d_{\mathrm{model}}$ using a $1\times1$ convolution (Conv1d). The output is then arranged as a sequence of latent vectors (tokens), each representing the estimated brain state over a short sub-interval, and passed to the temporal backbone.
\end{itemize}

\subsubsection{MAG Temporal Backbone}

After spatial tokenization, the feature sequence $X=\{x_1,x_2,\dots,x_T\}$ is processed by the temporal backbone. We use the Titans architecture, specifically the \textit{Memory-as-a-Gate (MAG)} variant, to combine efficient long-term memory with local attention-based modeling~\cite{titans_paper}.

\begin{figure*}[h]
    \centering
    \includegraphics[width=0.85\linewidth]{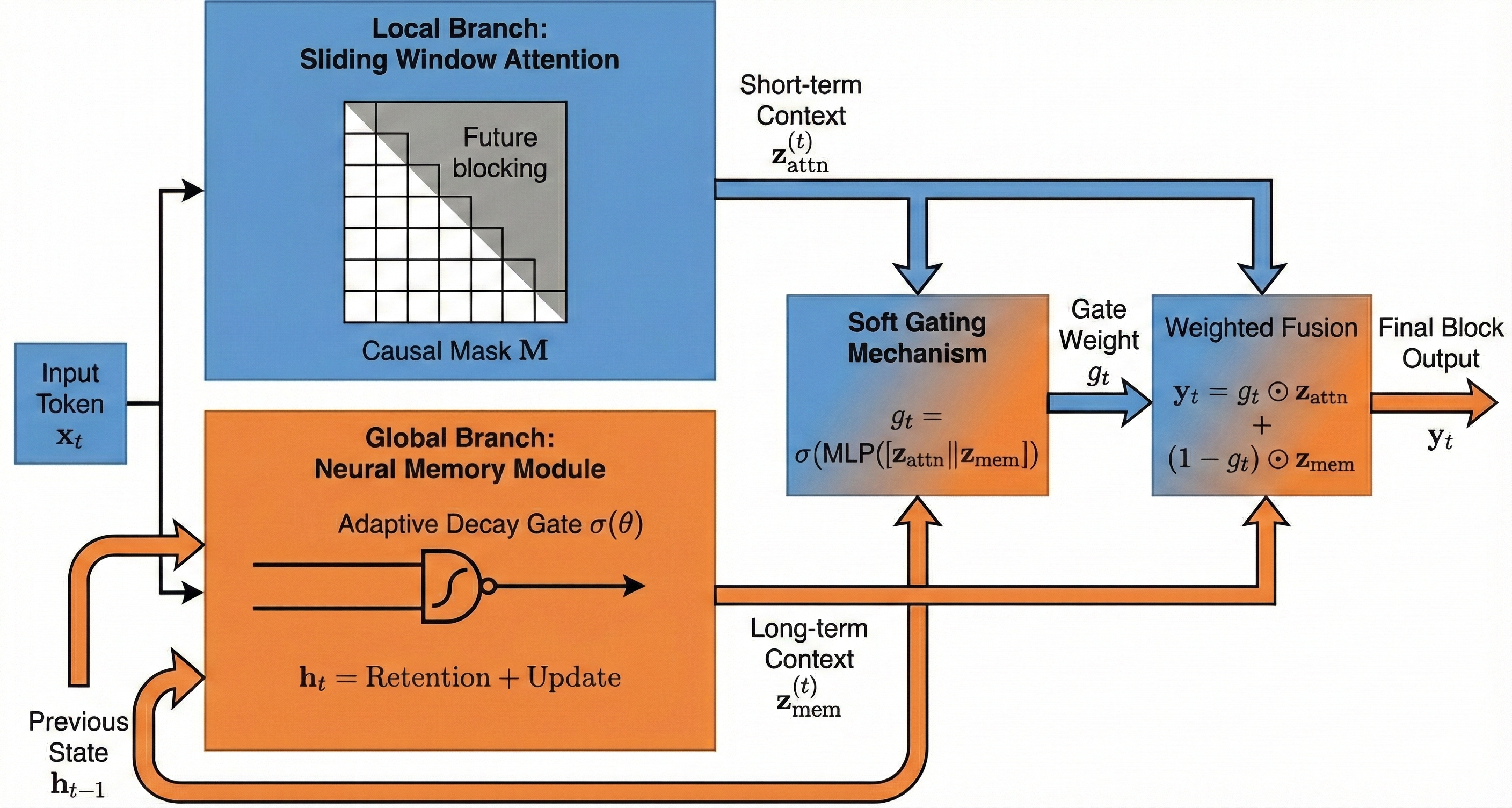}
    \caption{\textbf{Titans: Memory-as-a-Gate (MAG) module.} The module combines a \textbf{local branch} (causal sliding-window attention) that captures short-range patterns and a \textbf{global branch} (neural memory) that maintains long-range context. A learnable gate $g_t$ fuses the two representations into a context-aware output.}
    \label{fig:titans_block}
\end{figure*}

\paragraph{Rationale}
Seizure forecasting requires modeling both transient anomalies and longer-term pre-ictal evolution. Standard Transformers capture rich local interactions but are costly for long sequences due to the quadratic self-attention complexity $\mathcal{O}(N^2)$~\cite{vaswani2017attention}. Recurrent models such as LSTMs have linear complexity $\mathcal{O}(N)$ and are suitable for streaming settings, yet their effective ability to retain and retrieve information can degrade over very long horizons, making it difficult to link early pre-ictal changes to later seizure onset~\cite{hochreiter1997long}. Titans addresses this trade-off by augmenting attention with a compact neural memory that can store long-range information with fixed per-step computation~\cite{titans_paper}.

\paragraph{Neural Memory Module}
The neural memory maintains a persistent latent state $h_t$ updated at each time step via an adaptive decay mechanism~\cite{titans_paper}:
\begin{equation}
    \label{eq:neural_memory}
    h_t = \sigma(\theta) \odot h_{t-1} + \left(1 - \sigma(\theta)\right) \odot \mathcal{K}(x_t),
\end{equation}
where $h_{t-1}\in\mathbb{R}^{d_{\mathrm{model}}}$ is the previous memory state, $x_t$ is the current token, $\odot$ denotes element-wise multiplication, and $\sigma(\theta)$ is a learnable decay gate with parameters $\theta\in\mathbb{R}^{d_{\mathrm{model}}}$. Compared to fixed-rate exponential moving averages, the learnable $\theta$ allows the model to adjust retention per feature dimension, which can help attenuate transient artifacts while preserving informative pre-ictal trends over long horizons~\cite{mormann2007seizure}.

\paragraph{MAG Block}
At each time step $t$, MAG computes two complementary representations and fuses them via a data-dependent gate:
\begin{itemize}
    \item \textbf{Local branch (causal sliding-window attention):}
    The local branch models short-range dependencies using attention restricted to a causal neighborhood $\mathcal{N}_t$. To enforce autoregressive processing (no access to future steps), a causal mask $M\in\mathbb{R}^{S\times S}$ is applied:
    \begin{equation}
        M_{ij} =
        \begin{cases}
            0 & \text{if } j \le i, \\
            -\infty & \text{if } j > i.
        \end{cases}
    \end{equation}
    The attention output is:
    \begin{equation}
        z_{\mathrm{attn}}^{(t)} = \mathrm{Softmax}\!\left(\frac{Q_t K_{\mathcal{N}_t}^{\top}}{\sqrt{d_k}} + M \right)V_{\mathcal{N}_t}.
    \end{equation}
    This formulation prevents information leakage from future time steps during sequential simulation.

    \item \textbf{Global branch (neural memory):}
    In parallel, the global branch summarizes long-range context through the updated memory state $h_t$ (Eq.~\ref{eq:neural_memory}) and a linear projection:
    \begin{equation}
        z_{\mathrm{mem}}^{(t)} = W_{\mathrm{mem}} h_t + b_{\mathrm{mem}},
    \end{equation}
    where $W_{\mathrm{mem}}$ and $b_{\mathrm{mem}}$ are trainable parameters. This representation is intended to capture slowly varying background context over extended periods.

    \item \textbf{Soft gating fusion:}
    The two branches are fused using a learned gate:
    \begin{equation}
        g_t = \sigma\!\left(W_g [z_{\mathrm{attn}} \parallel z_{\mathrm{mem}}] + b_g\right),
    \end{equation}
    where $[\,\cdot \parallel \cdot\,]$ denotes concatenation. The fused output is:
    \begin{equation}
        y_t = g_t \odot z_{\mathrm{attn}} + (1 - g_t) \odot z_{\mathrm{mem}}.
    \end{equation}
    This gating mechanism allows the model to adaptively balance short-term patterns and long-term context depending on the current signal characteristics.
\end{itemize}

\subsubsection{Classification Head}
The final representation produced by the Titans backbone is passed to a linear classification layer to yield $p \in [0,1]$, representing the predicted probability of an upcoming seizure within the target horizon.

\subsection{Post-processing and Evaluation Strategy}
\label{sec:evaluation_strategy}

\subsubsection{Soft Fusion Top-K and Decision Rule}
To improve prediction stability in the presence of transient artifacts, we adopt a window-based soft decision scheme rather than applying an instantaneous hard threshold to each segment~\cite{lu2023cbam3dcnnlstm}. 

Let $P_W=\{p_{t-W+1},\dots,p_t\}$ denote the set of predicted probabilities within a sliding window of length $W$ ending at time step $t$. We compute an alarm score $S_t$ as the mean of the $K$ largest probabilities within $P_W$:
\begin{equation}
    S_t=\frac{1}{K}\sum_{p\in \mathrm{top}\text{-}K(P_W)} p.
\end{equation}
An alarm is triggered when $S_t>\tau$. Two additional mechanisms are used to support clinical practicality and reduce spurious alerts.

\begin{itemize}
    \item \textbf{Patient-specific threshold selection:}
    Because EEG amplitude distributions and baseline activity differ substantially across subjects, a single global threshold is often suboptimal. We therefore select $\tau$ separately for each subject by grid search on the validation set over $\tau\in[0.10,0.95]$ with step size 0.05. The selected $\tau$ maximizes sensitivity while keeping the false-positive rate within acceptable limits.

    \item \textbf{Refractory period:}
    To prevent repeated alarms triggered by a single prolonged event or artifact, we enforce a 30-minute refractory period~\cite{mormann2007seizure}. After an alarm occurs, subsequent threshold crossings within the next 30 minutes are ignored. This constraint ensures that a single episode contributes at most one alarm event, reducing redundant alerts and preventing inflation of false-positive counts.
\end{itemize}

\subsubsection{Evaluation Strategy: Sensitivity-Oriented Validation}
Many seizure prediction studies optimize a balance between sensitivity and false-positive rate using composite criteria (e.g., AUC or Youden index). In contrast, we adopt a sensitivity-oriented evaluation aligned with safety-first clinical use. Specifically, we treat high sensitivity as a prerequisite and then minimize false positives subject to that requirement.

\begin{itemize}
    \item \textbf{Sensitivity as a primary requirement:}
    Rather than optimizing sensitivity and false positives symmetrically, we aim to operate near maximal sensitivity and reduce the false-positive rate under this constraint. This formulation reflects the practical cost of missed seizures in safety-critical settings and discourages improvements driven primarily by conservative thresholding.

    \item \textbf{Assessment of long-context utilization:}
    Operating at very high sensitivity typically increases susceptibility to noise, which often manifests as elevated false-positive rates. If EEG-Titans maintains a low false-positive rate at such operating points, it provides supporting evidence that the neural memory branch contributes meaningful long-term context, enabling the model to differentiate progressive pre-ictal evolution from transient background fluctuations, rather than achieving performance mainly via threshold tuning.
\end{itemize}

\subsubsection{Quantitative Metrics}
We adopt the standard forecasting definitions based on two clinically motivated parameters~\cite{mormann2007seizure}: the \textbf{Seizure Prediction Horizon (SPH)} and the \textbf{Seizure Occurrence Period (SOP)}. We set $\text{SPH}=5$ minutes as the interval immediately preceding seizure onset during which an effective response is unlikely, and $\text{SOP}=30$ minutes as the time window after an alarm within which a seizure is expected. Accordingly, the target pre-ictal interval is defined as $[T_{\mathrm{onset}}-35,\,T_{\mathrm{onset}}-5]$ minutes.

\begin{itemize}
    \item \textbf{Segment-based sensitivity:}
    Seizure forecasting performance is commonly evaluated using event-based or segment-based criteria. Event-based evaluation counts a seizure as correctly predicted if at least one alarm occurs within the SOP, whereas segment-based evaluation assesses the classification of each segment independently. In this study, we report segment-based sensitivity to quantify whether the model consistently recognizes pre-ictal activity across the entire pre-ictal interval:
    \begin{equation}
        \mathrm{Sensitivity} = \frac{TP_{\mathrm{seg}}}{TP_{\mathrm{seg}} + FN_{\mathrm{seg}}}.
    \end{equation}
    Here, $TP_{\mathrm{seg}}$ denotes the number of segments within the pre-ictal interval correctly classified as pre-ictal, and $FN_{\mathrm{seg}}$ denotes the number of pre-ictal segments misclassified as inter-ictal. High segment-based sensitivity indicates sustained warning coverage rather than sporadic detections.

    \item \textbf{False positive rate per hour (FPR/h):}
    We measure usability using the false positive rate per hour, computed over inter-ictal data:
    \begin{equation}
        \mathrm{FPR/h} = \frac{N_{\mathrm{FP}}}{T_{\mathrm{inter\text{-}ictal}}},
    \end{equation}
    where $T_{\mathrm{inter\text{-}ictal}}$ is the total inter-ictal duration in hours and $N_{\mathrm{FP}}$ is the number of false alarm events. To reflect user experience and avoid double-counting when sustained artifacts produce repeated threshold crossings, false alarms occurring within the 30-minute refractory period are merged and counted as a single event~\cite{mormann2007seizure}. Maintaining a low FPR/h is essential to reduce alarm fatigue and support long-term adherence to a forecasting system.
\end{itemize}

\section{Results}
\label{sec:results_analysis}

Table~\ref{tab:main_results} summarizes the forecasting performance of EEG-Titans on the test set. Consistent with our sensitivity-oriented evaluation strategy, we analyze results by separating subjects with stable behavior from a small number of challenging cases that motivate patient-specific adaptation.

\begin{table}[h]
    \centering
    \caption{Baseline forecasting performance. The ``standard 60\,s context'' follows windowing conventions commonly adopted in prior benchmarks~\cite{truong2018generalised}. Shaded rows indicate subjects with elevated false positive rates before adaptation.}
    \label{tab:main_results}
    \renewcommand{\arraystretch}{1.2}
    \setlength{\tabcolsep}{12pt}
    \begin{tabular}{ccc}
        \toprule
        \textbf{Subject ID} & \textbf{Sensitivity (\%)} & \textbf{FPR/h} \\
        \midrule
        CHB01 & 100.00 & 0.0618 \\
        CHB02 & 100.00 & 0.0686 \\
        CHB03 & 100.00 & \textbf{0.0000} \\
        CHB04 & 100.00 & \textbf{0.0000} \\
        CHB05 & 100.00 & 0.3582 \\
        \rowcolor{lightred} CHB06 & 100.00 & 0.9034 \\
        CHB07 & 100.00 & 0.1846 \\
        CHB08 & 100.00 & 0.1303 \\
        CHB09 & 100.00 & 0.0834 \\
        CHB10 & 100.00 & \textbf{0.0000} \\
        CHB13 & 100.00 & 0.0663 \\
        CHB14 & 100.00 & 0.0983 \\
        \rowcolor{lightred} CHB15 & 90.34  & 3.8710 \\
        CHB16 & 100.00 & 0.1229 \\
        CHB18 & 100.00 & 0.5000 \\
        CHB20 & 100.00 & 0.0831 \\
        CHB21 & 100.00 & \textbf{0.0000} \\
        CHB22 & 100.00 & 0.1527 \\
        \midrule
        \textbf{Baseline Average} & \textbf{99.46} & \textbf{0.3713} \\
        \bottomrule
    \end{tabular}
\end{table}

\subsection{Stable Cohort}
For the majority of subjects (16 out of 18; 89\% of the cohort), EEG-Titans achieves 100\% segment-level sensitivity while maintaining a low false positive rate (typically $<0.2$ alarms/hour). This consistent behavior suggests that, for standard-quality recordings, the model learns a stable and discriminative mapping between pre-ictal dynamics and inter-ictal baseline activity. In particular, the Titans backbone appears able to leverage long-context information to reduce sensitivity to transient fluctuations while preserving high detection coverage, in line with the intended role of the neural memory pathway~\cite{titans_paper}.

\subsection{Adaptive Capability and Outlier Analysis}
\label{sec:outliers}

Although performance is stable for most subjects, two cases-CHB15 and CHB06-exhibit elevated false positive rates under the baseline configuration. Rather than treating these as purely statistical anomalies, we interpret them as clinically relevant edge cases that highlight the need for patient-specific adaptation strategies.

\paragraph{Demonstrated Adaptation}
CHB15 shows the highest baseline false positive rate (3.87/h), which is consistent with dense high-amplitude muscle artifacts in the recordings. Under the baseline configuration using a 60-second context (consistent with common protocols~\cite{truong2018generalised}), the model is more likely to confuse recurrent artifact patterns with sustained pre-ictal energy accumulation.

To examine the model’s adaptability, we apply the hierarchical input strategy by extending the context window from 60\,s to 300\,s (5 minutes). As reported in Table~\ref{tab:chb15_adaptation}, this adjustment yields a marked improvement.

\begin{table*}[h]
    \centering
    \caption{Effect of context extension for subject CHB15.}
    \label{tab:chb15_adaptation}
    \setlength{\tabcolsep}{10pt}
    \renewcommand{\arraystretch}{1.1}
    \begin{tabular}{lccc}
        \toprule
        \textbf{Setting} & \textbf{Context Window} & \textbf{Sen (\%)} & \textbf{FPR/h} \\
        \midrule
        Baseline & 60\,s (1\,min) & 90.34 & 3.87 \\
        \textbf{Adaptive (New)} & \textbf{300\,s (5\,min)} & \textbf{99.86} & \textbf{0.00} \\
        \bottomrule
    \end{tabular}
\end{table*}

\textit{Analysis:} The improvement to 0.00 FPR/h indicates that extending the local context can substantially enhance robustness under heavy artifact contamination. A plausible interpretation is that a wider sliding-window attention span enables the local branch to model temporal consistency over several minutes and to down-weight discontinuous artifact bursts that lack coherent evolution. In turn, this reduces the likelihood that short-lived artifacts are propagated into the global memory pathway, improving both specificity and sensitivity in this subject.

\paragraph{Potential vs.\ Integrity}
CHB06 achieves 100\% sensitivity but exhibits an elevated false positive rate (0.90/h). This pattern likely reflects subject-specific and demographic factors in the dataset. First, CHB06 is the youngest subject in the cohort (approximately 1.5 years old), and pediatric EEG can be highly non-stationary with developmental dynamics that differ substantially from older children and adults. Second, the seizures in this subject are relatively brief (on the order of $\sim$15 seconds), which can bias the operating point toward higher sensitivity and thereby increase susceptibility to spurious fluctuations.

While additional calibration or adaptation (e.g., frequency-weighted filtering or subject-specific regularization) may reduce false alarms in CHB06, we intentionally report the baseline performance to reflect the genuine heterogeneity present in CHB-MIT. This case motivates the need for a dedicated pediatric calibration protocol in future deployments, rather than relying on retrospective optimization that may mask clinically important variability.


\subsection{Benchmarking Against Prior CHB-MIT Results}
\label{sec:comparison}

Table~\ref{tab:comparison} positions EEG-Titans (standard configuration) against representative state-of-the-art (SOTA) methods reported on the CHB-MIT dataset.

\begin{table*}[h]
    \centering
    \caption{Performance comparison with state-of-the-art methods on the CHB-MIT dataset.}
    \label{tab:comparison}
    \setlength{\tabcolsep}{6pt}
    \renewcommand{\arraystretch}{1.15}
    \begin{tabular}{l p{4.2cm} c p{5.5cm} c c}
        \toprule
        \textbf{Category} & \textbf{Study / Model} & \textbf{Year} & \textbf{Methodology} & \textbf{Sen (\%)} & \textbf{FPR/h} \\
        \midrule
        \multicolumn{6}{l}{\textit{\textbf{Transformer-based}}} \\
        & ~\cite{yan2022three} & 2022 & 3-Tower Transformer + STFT & 96.01 & 0.047 \\
        & ~\cite{hu2023hybrid} & 2023 & Hybrid Transformer + Wavelet & 91.70 & 0.000 \\
        & ~\cite{zheng2024two} & 2024 & Two-stage Set Transformer & 80.10 & 0.110 \\
        & ~\cite{hussein2022multi} & 2022 & Multi-channel ViT + CWT & 99.80 & 0.004 \\
        \midrule
        \multicolumn{6}{l}{\textit{\textbf{TCN \& Hybrid}}} \\
        & ~\cite{huang2024tcn} & 2024 & TCN + Self-Attention & 97.37 & N/A \\
        & ~\cite{lu2023cbam3dcnnlstm} & 2023 & CBAM-3D CNN + BiLSTM (STFT) & 98.40 & 0.017 \\
        \midrule
        \multicolumn{6}{l}{\textit{\textbf{Proposed Approach}}} \\
        & \textbf{Ours (Memory)} & \textbf{2025} & \textbf{Titans + Sliding Attention} & \textbf{99.46} & \textbf{0.371} \\
        \bottomrule
    \end{tabular}
\end{table*}

Several recent methods report sensitivities above 99.8\% (e.g.~\cite{hussein2022multi}). Under the baseline setting, EEG-Titans attains an average sensitivity of 99.46\%, indicating that the proposed memory-augmented backbone is competitive with strong CNN and Transformer baselines. Notably, multiple SOTA approaches rely on explicit time--frequency transformations (e.g., STFT or CWT), whereas EEG-Titans operates on tokens extracted from minimally processed EEG via the ShallowConvNet tokenizer. This suggests that learned spectro-spatial tokenization combined with memory-augmented temporal modeling can provide sufficient representational capacity for high-coverage seizure forecasting in this setting.

Within the hybrid category, EEG-Titans substantially improves upon a recent CNN+LSTM pipeline~\cite{lu2023cbam3dcnnlstm}, increasing sensitivity from 98.40\% to 99.46\%. This gain is consistent with the hypothesis that the \textit{Memory-as-a-Gate} mechanism preserves and integrates long-range information more effectively than standard recurrent units, whose effective memory can degrade over extended horizons~\cite{hochreiter1997long, titans_paper}.

EEG-Titans yields a baseline of 0.371 FPR/h. While this value is higher than the ultra-low FPR/h reported by some efficiency-oriented CNN approaches, it should be interpreted in the context of our sensitivity-oriented threshold selection and cohort heterogeneity. A non-trivial fraction of false alarms is concentrated in a small number of challenging subjects (notably CHB15 and CHB06), making the cohort-average FPR/h sensitive to artifact burden and demographic variability. Importantly, the CHB15 analysis shows that subject-specific adaptation through context extension can reduce false alarms without compromising sensitivity, indicating a practical pathway for improving FPR/h beyond the baseline configuration.

\subsection{Ablation Study}
\label{sec:architectural_insights}

To better attribute performance to architectural components, we compare EEG-Titans with simplified temporal backbones. Table~\ref{tab:ablation_average} reports average results across 18 subjects.

\begin{table*}[htbp]
    \centering
    \caption{Ablation study: average performance across 18 subjects.}
    \label{tab:ablation_average}
    \renewcommand{\arraystretch}{1.2}
    \setlength{\tabcolsep}{5pt}
    \begin{tabular}{lcc}
        \toprule
        \textbf{Model Configuration} & \textbf{Avg. Sensitivity (\%)} & \textbf{Avg. FPR/h} \\
        \midrule
        \textbf{Proposed (EEG-Titans)} & \textbf{99.46} & \textbf{0.371} \\
        \midrule
        Baseline 1: LSTM (Temporal Only) & 98.70 & 0.395 \\
        Baseline 2: Attention (Spatial Only) & 98.41 & 0.522 \\
        \bottomrule
    \end{tabular}
\end{table*}

\subsubsection{Tokenizer-driven separability across temporal backbones}
All variants achieve high sensitivity ($>98\%$), indicating that the ShallowConvNet-based \textbf{spatial tokenizer} produces discriminative spectro-spatial representations~\cite{schirrmeister2017deep}. This behavior is consistent with the tokenizer acting as a learnable filtering stage that emphasizes informative power patterns and spatial projections commonly used in EEG decoding. Under this interpretation, the temporal backbone primarily enforces temporal consistency and helps reject spurious activations rather than discovering seizure-related features from scratch.

\subsubsection{Neural memory reduces false alarms under the same tokenizer}
While sensitivity is high across baselines, EEG-Titans achieves the lowest FPR/h, improving upon the attention-only temporal baseline (0.371 vs.\ 0.522). This reduction supports the role of the \textbf{Memory-as-a-Gate} design in suppressing isolated high-confidence predictions that are inconsistent with longer-term context. The dual-branch structure maintains responsiveness through the local attention pathway while using the memory pathway as a slower-varying contextual reference to down-weight transient deviations.


A practical question is whether EEG-Titans's gains arise from the memory mechanism itself or simply from increased model capacity. Although parameter-matched controls would be required for definitive attribution, the observed results and architectural properties are consistent with the neural memory pathway providing functional benefits beyond passive capacity scaling.

\subsubsection{Long-horizon retention: Titans memory versus LSTM recurrence}
LSTMs reduce vanishing gradients relative to earlier RNNs~\cite{hochreiter1997long}, but their effective ability to retain and retrieve information can still weaken as sequence length increases. Titans maintains a persistent memory state updated by an adaptive retention mechanism~\cite{titans_paper}, designed to preserve salient information over long horizons with fixed per-step computation. The consistent FPR/h reduction relative to the LSTM temporal baseline (Table~\ref{tab:ablation_average}) is compatible with the view that neural memory provides a more reliable long-context reference for rejecting transient artifact-driven activations.

\subsubsection{Selective context gating rather than passive smoothing}
A concern is that neural memory might act mainly as a smoothing operator. Pure smoothing would typically suppress rapid changes and could increase false negatives for brief events. In CHB06, which contains very short seizures ($\sim$15\,s), EEG-Titans maintains 100\% sensitivity (Section~\ref{sec:results_analysis}), suggesting that the model does not simply flatten fast dynamics. Instead, the data-dependent gating in the MAG block provides a plausible mechanism for selectively amplifying informative short-term patterns while using longer-term context to suppress non-coherent fluctuations~\cite{titans_paper}, consistent with an active, context-conditioned memory rather than passive smoothing.

\section{Discussion}
\label{sec:discussion}

This work investigates whether modern neural memory can improve EEG-based seizure forecasting under long-horizon, sequential evaluation, where subtle pre-ictal changes must be detected without violating temporal causality. Across 18 CHB-MIT subjects, EEG-Titans achieves high segment-level sensitivity under a chronological hold-out protocol, while revealing interpretable failure modes in artifact-heavy and highly non-stationary recordings. Below we discuss the implications of these findings for long-context modeling, clinical deployment, and future extensions.

\subsection{Clinical perspective: sensitivity-first operation under causal evaluation}
Seizure forecasting is a safety-critical task in which missed events can have severe consequences for patients; thus, it is clinically reasonable to treat high sensitivity as a primary constraint rather than a tunable trade-off~\cite{mormann2007seizure,cook2013prediction}. In contrast to retrospective protocols that risk temporal leakage, our chronological hold-out design better reflects the real deployment scenario in which the model must generalize from earlier history to future seizure clusters~\cite{truong2018generalised}. Importantly, this sequential setting makes false alarm control more challenging, because operating near maximal sensitivity typically amplifies the impact of non-stationarity and artifacts. The observed concentration of false alarms in a small subset of subjects reinforces the view that patient heterogeneity and recording conditions dominate practical performance in long-term EEG monitoring~\cite{shoeb2009application,mormann2007seizure}.

\subsection{Why neural memory matters for long-horizon pre-ictal dynamics}
A core motivation of EEG-Titans is that pre-ictal biomarkers are often subtle, non-stationary, and unfold gradually over extended durations, requiring models to integrate information across distant time points~\cite{truong2018generalised,mormann2007seizure}. Classical recurrent units, including LSTMs, can degrade in effective memory as sequence length grows, making it difficult to reliably link early pre-ictal changes to later seizure onset under long horizons~\cite{hochreiter1997long}. Transformers alleviate local context limitations but incur quadratic cost with sequence length, which becomes prohibitive for very long windows and resource-constrained devices~\cite{vaswani2017attention}. EEG-Titans leverages the Titans Memory-as-a-Gate design to maintain a persistent long-context state with fixed per-step computation, enabling the model to preserve salient long-range information while retaining responsiveness to local anomalies. The ablation results support this motivation: when the tokenizer is held constant, the memory-augmented temporal backbone reduces FPR/h relative to attention-only and LSTM-based baselines, suggesting that memory provides a stabilizing reference for rejecting transient, non-coherent activations.

\subsection{Tokenizer--backbone synergy and EEG-specific inductive bias}
All temporal variants achieve high sensitivity, indicating that the ShallowConvNet tokenizer provides strong spectro-spatial representations from minimally processed EEG. This aligns with established EEG decoding practice in which spatial filtering and power-related features capture discriminative neural signatures~\cite{bandarabadi2015epileptic,schirrmeister2017deep}. Under this view, the temporal backbone’s primary role shifts from feature discovery to enforcing temporal consistency and artifact rejection. The MAG fusion in EEG-Titans is well matched to this requirement: the local attention stream can react quickly to candidate signatures, while the memory stream contextualizes those signatures against longer-term trends, reducing spurious alarms when local evidence is not supported by sustained evolution~\cite{titans_paper}.

\subsection{Subject heterogeneity and context as a control knob for artifact rejection}
The CHB-MIT cohort includes substantial variability in age, seizure morphology, and recording artifacts~\cite{shoeb2009application}. Two observations are particularly informative. First, the CHB15 case suggests that false alarms driven by muscle artifacts can be suppressed by extending temporal context, which enables the model to discriminate temporally incoherent bursts from progressively evolving pre-ictal trajectories. Second, the CHB06 case highlights biological non-stationarity in very young subjects, where brief seizures and developmental EEG dynamics can induce hypersensitivity and elevated false positives even when sensitivity remains high. Together, these results support a practical conclusion: temporal context size is a clinically meaningful tuning axis for balancing robustness and responsiveness under safety-first operation. An important next step is to replace manual adaptation with an automated policy (e.g., uncertainty- or artifact-aware context selection) that can adjust context length online without violating causality~\cite{mormann2007seizure,truong2018generalised}.

\subsection{Long-sequence modeling, early anticipation, and trustworthy deployment}
\label{sec:discussion_broader_context}

EEG-Titans relates to three converging directions in modern EEG modeling: (i) efficient long-sequence representation, (ii) early anticipation under weak and slowly evolving biomarkers, and (iii) trustworthy deployment under strong subject heterogeneity and safety constraints. We discuss each aspect below and position the present study within the broader literature.

\subsubsection{Efficient long-horizon sequence modeling in EEG}
A central challenge in seizure forecasting is that pre-ictal signatures may unfold over tens of minutes and can be interleaved with non-stationary artifacts, requiring models to integrate information over long horizons while remaining computationally feasible~\cite{mormann2007seizure,truong2018generalised}. Classical recurrent architectures (e.g., LSTMs) were introduced to mitigate vanishing gradients, yet their effective memory can still degrade on ultra-long sequences in practice~\cite{hochreiter1997long}. Transformer-style self-attention improves global context modeling, but its quadratic cost with sequence length constrains scalability for long observation windows and resource-limited deployment~\cite{vaswani2017attention}. As a result, several recent seizure prediction pipelines rely on limited context windows and/or explicit time--frequency transformations (e.g., STFT/CWT) coupled with heavier backbones~\cite{yan2022three,hussein2022multi}.

Beyond attention and recurrence, efficient sequence modeling families such as state-space models (SSMs) have recently gained traction for long-context learning due to their favorable scaling and stable dynamics \cite{gu2024mamba}. In EEG/ECG applications, SSM-based designs have been explored for clinical classification and representation learning under long recordings, emphasizing efficient long-range temporal modeling and robustness to temporal variability~\cite{tran2024eeg, nguyen2025ecg}. In this context, EEG-Titans provides an alternative route to long-horizon modeling by using a neural memory mechanism with fixed per-step computation. Conceptually, memory-augmented models and SSMs share the objective of scalable long-context processing, suggesting that future EEG forecasting systems may benefit from hybrid designs that combine structured latent dynamics with adaptive memory gating.

\subsubsection{Early anticipation as a shared paradigm across EEG applications}
Seizure forecasting is fundamentally an \emph{anticipation} problem: the system must recognize subtle, gradually evolving dynamics before the overt ictal transition, under strict temporal causality and with limited tolerance for missed events~\cite{mormann2007seizure,cook2013prediction}. This anticipatory framing is shared by other EEG tasks where discriminative neural signatures emerge before observable outcomes. For example, studies on perceptual decision-making have shown that EEG correlates of evidence accumulation (e.g., the centroparietal positivity) evolve continuously prior to overt responses, supporting the need for models that integrate information over time rather than rely on isolated window-level snapshots~\cite{tran2023early,oconnell2012supramodal,kelly2013jneurosci,tagliabue2019eeg}. Similarly, work on post-choice dynamics suggests that decision-related neural variables can continue to evolve after initial commitment, reinforcing the value of architectures that maintain and update context across extended horizons~\cite{murphy2015elife}. In a related vein, multimodal forecasting of human decision performance highlights that anticipatory cues are often progressive and benefit from temporal integration~\cite{tran2024multimodal}.

From this viewpoint, EEG-Titans’s dual-branch design reflects a general anticipation strategy: a responsive pathway that detects local anomalies, coupled with a stabilizing pathway that integrates longer-term context to validate whether local deviations are consistent with a true state transition. This perspective is consistent with prior seizure prediction findings that models capable of accumulating evidence over longer horizons can be more resilient than approaches operating on short independent segments.

\subsubsection{Trustworthy forecasting: interpretability, fairness, and patient-level reliability}
Real-world deployment of seizure forecasting systems depends not only on average performance but also on trust: clinicians and users must understand when and why alarms occur, and the system must behave reliably across heterogeneous patients and recording conditions~\cite{shoeb2009application,mormann2007seizure}. The concentration of false alarms in artifact-heavy or highly non-stationary subjects underscores that cohort-level averages can mask clinically important failure modes. This motivates evaluation beyond aggregate metrics, including subject-level stratification and safety-first operating points, as well as explicit analyses of challenging subgroups.

In parallel clinical domains, fairness-aware evaluation frameworks have been proposed to assess whether model behavior remains consistent across demographic strata and recording variations, particularly when explainability is used to support clinical plausibility checks~\cite{nguyen2024fairad}. Although originally studied in Alzheimer’s disease detection, the same principle applies to seizure forecasting: systematic over-alarming in specific subgroups (e.g., very young subjects or recordings with frequent muscle artifacts) may be unacceptable despite strong mean sensitivity. Furthermore, contrastive learning has been used to improve EEG representation robustness by promoting invariant structure across conditions or modalities, which can improve calibration under distribution shifts and reduce spurious activations in noisy settings~\cite{tran2024eeg}.


\subsection{Limitations and future directions}
Several limitations should be acknowledged. First, CHB-MIT is a pediatric cohort with specific hospital recording characteristics; external validation on additional datasets and prospective evaluation are needed to assess generalization in ambulatory settings. Second, although the chronological protocol mitigates leakage, patient-specific thresholding and context adaptation remain important; future work should formalize calibration policies that are efficient and clinically interpretable, potentially guided by artifact detection and uncertainty estimation. Third, interpretability is not explicitly addressed in the current model; incorporating explanation tools (e.g. token-/channel-level attribution aligned with seizure onset zones) and evaluating fairness across age/sex strata would strengthen clinical readiness. Finally, while EEG-Titans uses minimally processed signals, integrating complementary modalities (e.g., ECG or wearable context signals) and leveraging representation learning (e.g., contrastive pretraining) may further improve robustness under real-world noise and non-stationarity.

\section{Conclusion}
\label{sec:conclusion}

This work presented EEG-Titans, a seizure prediction framework that combines ShallowConvNet tokenization with a Titans Memory-as-a-Gate backbone to model long-range pre-ictal dynamics while remaining sensitive to short-term abnormalities. On CHB-MIT with chronological hold-out evaluation, the model achieved 99.46\% average segment-level sensitivity with 0.371 FPR/h, and showed stable behavior for 16/18 subjects (100\% sensitivity with low false-alarm rates). The two challenging cases (CHB15, CHB06) highlight the impact of artifacts and pediatric non-stationarity under a sensitivity-first operating point. Importantly, CHB15 demonstrates that false-alarm control is tightly linked to temporal context - extending the context from 60\,s to 300\,s reduced FPR/h from 3.87 to 0.00 while increasing sensitivity from 90.34\% to 99.86\%. These results support neural-memory-based backbones as a practical approach for long-context EEG forecasting and motivate future work on patient-specific calibration for robust deployment.

\bibliographystyle{named}
\balance
\bibliography{ijcai19.bib}

\end{document}